\documentclass[letterpaper, 10 pt, conference]{ieeeconf}  

\IEEEoverridecommandlockouts                              

\usepackage{lipsum}
\usepackage{graphicx}
\usepackage[acronym]{glossaries}
\usepackage[disable]{todonotes}
\setuptodonotes{inline}
\usepackage{amsmath,amsfonts,amssymb}
\usepackage{verbatim}
\usepackage{bm}
\usepackage{caption}
\usepackage{subcaption}
\usepackage[
backend=bibtex,
style=numeric,
sorting=none
]{biblatex}
\usepackage[english]{babel}

\newdimen\figrasterwd
\newacronym{rl}{RL}{Reinforcement Learning}
\newacronym{drl}{DRL}{Deep Reinforcement Learning}
\newacronym{dnn}{DNN}{Deep Neural Network}
\newacronym{nn}{NN}{Neural Network}
\newacronym{tl}{TL}{Transfer Learning}
\newacronym{asvgd}{ASVGD}{Amortized Stein Variational Gradient Descent}
\newacronym{mdp}{MDP}{Markov Decision Proces}

\title{\LARGE \bf
Towards Task-Prioritized Policy Composition
\todo[color=green]{Here we need a more general title. e.g.\: 
Towards combining policies with priorities in reinforcement learning
}
}

\author{Finn Rietz$^{1}$, Erik Schaffernicht$^{1}$, Todor Stoyanov$^{1,2}$ and Johannes A. Stork$^{1}$
\thanks{$^{1}$Autonomous Mobile Manipulation Lab, Center for Applied Autonomous Sensor Systems (AASS), Örebro University, Sweden.}%
\thanks{$^{2}$Department of Computing and Software, McMaster University, Canada. Correspondence: {\tt\small finn.rietz@oru.se}. This work was partially supported by the Wallenberg AI, Autonomous Systems and Software Program (WASP) funded by the Knut and Alice Wallenberg Foundation.}}

\begin{document}

\maketitle
\thispagestyle{empty}
\pagestyle{empty}

\begin{abstract}
Combining learned policies in a prioritized, ordered manner is desirable because it allows for modular design and facilitates data reuse through knowledge transfer.
In control theory, prioritized composition is realized by null-space control, where low-priority control actions are projected into the null-space of high-priority control actions. Such a method is currently unavailable for Reinforcement Learning. 
\todo[color=green]{Tell what this paper is about}
\todo[color=green]{Describe how it solves the problem} 
We propose a novel, task-prioritized composition framework for Reinforcement Learning, which involves a novel concept: The ``indifferent-space'' of Reinforcement Learning policies.
\todo[color=green]{Emphasize what is new or better}
Our framework has the potential to facilitate knowledge transfer and modular design while greatly increasing data efficiency and data reuse for Reinforcement Learning agents. Further, our approach can ensure high-priority constraint satisfaction, which makes it promising for learning in safety-critical domains like robotics. Unlike null-space control, our approach allows learning globally optimal policies for the compound task by online learning in the indifference-space of higher-level policies after initial compound policy construction.
\end{abstract}

\section{Introduction}

\todo[color=green]{Start with motivation as to why we want to combine policies. What are the benefits (e.g.\ transfer of knowledge and the implied increase of sample efficiency, modularity, exploiting that we already know to do certain tasks, etc.)}
Policy composition, i.e.\ the combination of already learned policies into a more capable compound policy, is desirable in \acrfull{rl}. Composition also increases modularity and allows for knowledge transfer between tasks, thus enabling efficient use of data and learning time. 
\todo[color=green]{Discuss that it is unclear what combining policies actually means. Discuss that you can look at this as doing things at the same time (as in adding the rewards) or as ordering tasks and doing them when they do not conflict (as done in control). Both approaches are valid, but they are different. There are some results for the ¨first approach, but they have shortcomings and they do not model the case when both task cannot be done at the same time.}
State-of-the-art \acrshort{rl} literature offers two perspectives on composition: Conjunctive (``AND") composition~\cite{russell2003q, haarnoja2018composable, hunt2019composing}, where tasks have to be solved concurrently, and disjunctive (``XOR") composition~\cite{precup1998theoretical}, where tasks are switched over time. When the individual tasks are very different, ``AND" composition often leads to sub-optimal policies and ``XOR" composition always ignores some of the tasks.
Classic control literature provides yet another perspective on composition:
Hierarchical task composition, where a low-priority control action is projected into the null-space of a high-priority control action such that the high-priority task is still achieved by the combined action~\cite{nakamura1987task}. 
This method, implemented by null-space control, is a form of local, prioritized task composition that is re-computed at each step. However, this can result in ``deadlocks'' (local optima) where the agent fails to achieve low-priority tasks in the long term.

\begin{figure}
\hfill
\includegraphics[height=0.45\hsize]{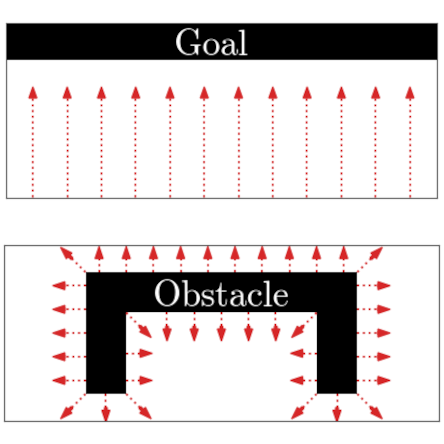}
\hfill
\includegraphics[height=0.435\hsize]{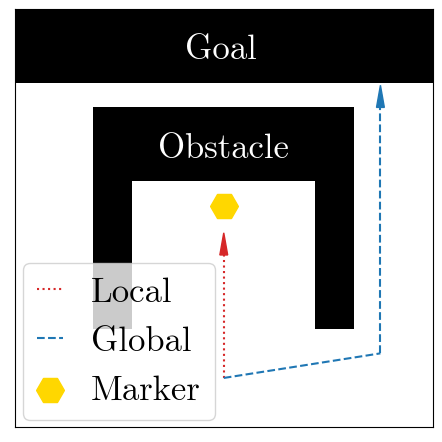}
\hfill
\caption{%
Optimal policies trained in separate environments for goal reaching (top left) and obstacle avoidance (bottom left).
Target environment with both tasks (right). Resulting behavior from local prioritized policy composition and globally optimal policy indicated by arrows.
}
\label{fig:goal_obstacle_policy}
\label{fig:env}
\end{figure}

\todo[color=green]{Discuss what are the benefits of ordering task or giving them a priority and when can that be useful (e.g.\ safety tasks). It is a common approach in control.}
In this paper, we are interested in task-prioritized composition similar to null-space control but for \acrshort{rl}. 
Consider the scenario in Fig.~\ref{fig:goal_obstacle_policy}, where we first train two \acrshort{rl} policies for two separate tasks and want to combine them for a new, more complex task. 
The first task (modeled by reward) $r_1$ is to avoid the U-shaped obstacle and the second task $r_2$ is to reach the goal. The arrows in Fig.~\ref{fig:goal_obstacle_policy} (left) show optimal policies $\pi_1^*$ and $\pi_2^*$ for the two tasks. The two tasks are completely separate such that $\pi_1^*$ never has to reach to the goal and $\pi_2^*$ never interacts with the obstacle during training. In the new task (right), we want to reach the goal while avoiding the obstacle and for this, we want to combine already learned policies $\pi^*_1$ and $\pi^*_2$ to a new policy.


\todo[color=green]{Explain that ordering tasks cannot be done by just adding reward (e.g.\ with an example) and that all approaches that are based on adding up rewards are therefore not sufficient! The robot will crash into the obstacle.}

Existing ``AND" composition approaches model the new task as the sum~\cite{russell2003q}, the mean~\cite{haarnoja2018composable}, or a convex combination of the reward signals~\cite{hunt2019composing}: $r_b = b r_1 + (1-b) r_2$, $b \in [0, 1]$.
\todo[color=green]{Give concrete example, describe how, because of this and depending on the reward scale for both signals, the agent might greedily drive into the obstacle with the new three panel figure.}
However, none of these approaches are suited for task-prioritized composition because they cannot implement a strict priority ordering of tasks. Even when we define the new task as $r = 0.9r_1 + 0.1 r_2$, existing methods cannot guarantee that the compound policy avoids the obstacle. Depending on the scale of individual rewards $r_1$ and $r_2$, the compound policy will drive straight into the obstacle.
What we need is a compound policy that follows $\pi^*_1$ to avoid the obstacle (with high priority) while making progress towards the goal of $\pi^*_2$ (with low priority) whenever possible. For this, the contribution of the low-priority policy has to be adjusted such that it does not interfere with the high-priority policy, e.g.\ in safety-critical situations where the high-priority policy must be in full control (for example at the marker in Fig.~\ref{fig:env}).
%
\todo[color=green]{Discuss that it is not clear what combining tasks with priorities means in RL, different to that we know what it means to do two tasks at the same time (which means adding the rewards). What is the equivalent formalization for combination with priorities?}

While ``AND" composition in \acrshort{rl} is well-defined as obtaining a policy a sum of rewards, e.g.\ $r_1 + r_2$, for task-prioritized policy composition it is not clear what the new task's reward is because the contribution of the low-propriety task depends on the policies of the high-priority task. Informally, we are interested in obtaining the policy $\pi_{1\succeq2}$ for a new task $r_{1\succeq2}$ that jointly optimizes a high-priority task $r_1$ and a low-priority task $r_2$ in such a way that the return for the high-priority task $r_1$ is not considerably lower under the composed policy $\pi_{1\succeq2}$ compared to the original, optimal policy $\pi_1^*$ for the task $r_1$. 
\todo{Maybe here we can remove ``additive composition is insufficient'' sentence, since this was already stated before, although for different reasons}
``AND" composition frameworks are insufficient for this, 
because the optimal policy for the compound task $r_{1+2}$ will very likely obtain less reward under $r_1$ compared to the optimal task policy $\pi_1^*$, due to the influence of the low-priority reward signal $r_2$~\cite{haarnoja2018composable, hunt2019composing}, a phenomenon referred to as ``tragedy of the commons''~\cite{russell2003q}.

More formally, we think of task-prioritized policy composition in  \acrshort{rl}  as solving a new task $r_{1\succeq2} = r_1 + w(s_t, a_t) r_2$, where $w(s_t, a_t) \in [0, 1]$ is some weight that is small whenever the contribution of the low-priority task $r_2$ would considerably worsen return for the high-priority task $r_1$ in the long run. This effectively prevents the low-priority task $r_2$ from contributing to the action-selection in such situations.
This means that our compound task $r_{1\succeq2}$ is only indirectly defined by the mentioned constraint on the return for $r_1$.

\todo[color=green]{Discuss that the intuitive notion is that the lower priority tasks should not reduce the return that the higher priority task can collect. Clearly, that means that we cannot just add the rewards. The reward of the lower priority task could influence the higher priority task's return. Just switching between the optimal policies of the two tasks will also be work.}

In the following, we first explain how policies with priorities are locally composed in null-space control, then we introduce the novel concept of ``indifference space" of \acrshort{rl} policies. After that, we explain how we propose to use the ``indifference space" to compose \acrshort{rl} policies that observe task priorities and how our method can obtain globally optimal solutions to prioritized task compositions by learning online. 



\begin{figure*}[!ht]
	\hspace{1cm}
	\begin{subfigure}[t]{0.6\textwidth}
		\includegraphics[width=\textwidth]{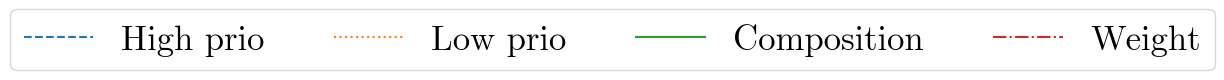}
	\end{subfigure}
	\par
	\vspace{-0.5cm}
	\centering
	\begin{subfigure}{0.3\textwidth}
		\centering
		\includegraphics[width=\textwidth]{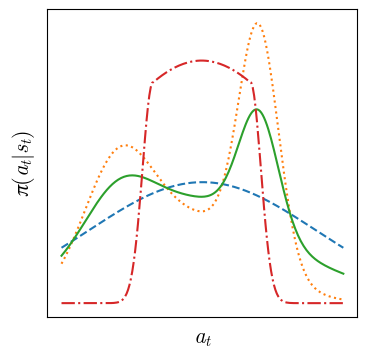}
		\caption{Permissive high-priority policy.}
		\label{fig:permissive_high_prio}
	\end{subfigure}
	\begin{subfigure}{0.3\textwidth}
		\centering
		\includegraphics[width=\textwidth]{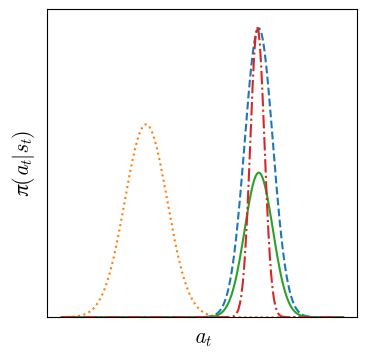}
		\caption{Restrictive high-priority policy}
		\label{fig:restrictive_high_prio}
	\end{subfigure}
	\begin{subfigure}{0.3\textwidth}
		\centering
		\includegraphics[width=\textwidth]{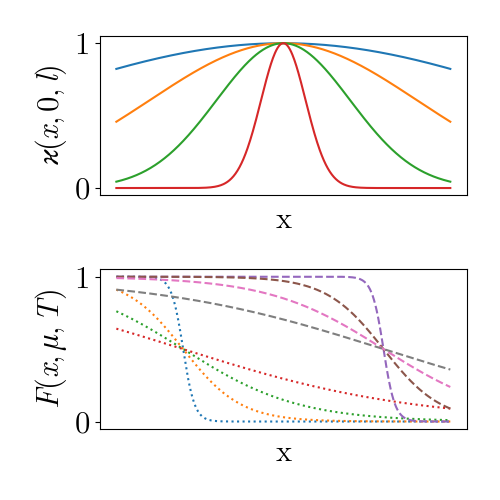}
		\caption{Top: SE kernel activations with varying length parameter $l$. Bottom: Fermi-Dirac distribution with varying $\mu$ and $T$.}
		\label{fig:weight_params}
	\end{subfigure}
	\caption{Desired weights as result of Fermi-Dirac and SE kernel parametrization in restrictive and permissive scenarios.}
	\label{fig:gauss_fermi_double_figure}
	
\end{figure*}

\section{Task composition in null-space control}
\label{sec:null-space-control}

\todo[color=green]{Discuss that there is a notion of combining tasks with priority in classical control: null-space control. Explain how it works and what it does in short terms.}
In control-theory, prioritized task combination is achieved through null-space control~\cite{nakamura1987task, siciliano2008robotics}. Null-space control involves a hierarchy of tasks and attempts to solve a lower-priority task in the null-space of a higher-priority task, which is that part of space that contains all equally valid solutions to the higher-priority task. Null-space control is possible when control vectors can be added together, in which case it is implemented by projecting the low-priority control vector $\bm{\theta}_2$ into the null-space of the high-priority controller. For this, we multiply $\bm{\theta}_2$ with a filtering term $(\mathbf{I}-\mathbf{J}^\intercal_{ee}\mathbf{J}^{\intercal+}_{ee})$, where $\mathbf{I}$ is the identity matrix, $\mathbf{J}^\intercal_{ee}$ is the transpose of the task Jacobian $\mathbf{J}_{ee}$ that relates changes in control with changes in task space, with $\mathbf{J}^{\intercal+}_{ee}$ being its pseudo inverse. This allows for the prioritized combination of a high-priority control signal $\bm{\theta}_1$ and a low-priority control signal $\bm{\theta}_2$ such that the resulting control $\bm{\theta}_{1\succeq2}$ 
will only execute those components of $\bm{\theta}_2$ that do no interfere with the desired task-space effect of $\bm{\theta}_1$.

\todo[color=green]{Discuss that this does not implement a globally optimal controller but ensures the priority locally because it does not deal with the long-term consequences of the local combination of actions.}
Importantly, null-space control does \emph{not} guarantee that the control $\bm{\theta}_{1\succeq2}$ will produce globally optimal behavior, i.e.\ it is possible that only the high-priority control will be achieved (i.e.\ collision avoidance), 
while the null-space projection of the low-priority control results in a global sub-optimal outcome. In Fig.~\ref{fig:env}, the robot would drive in a straight line towards the goal (in the null-space of the high-priority controller) up to the marker, at which point the high-priority controller would prevent it from advancing further to prevent a collision with the obstacle. 
Thus, classic null-space control can be thought of as a \textit{local} task-prioritized policy composition because the null-space projection is applied separately in every state, without considering the long-term consequences of this projection.

\section{Indifference-spaces of policies}
\todo[color=yellow]{Discuss how the local priority combination approach from control relates to RL, e.g.\ the controllers and their error error functions and in RL we have policies and their Q functions.
}

\todo[color=green]{Discuss that we can adopt the notion of null-space for error functions to RL by looking at null-spaces of Q functions. That is in the case that the policy allows for several optimal actions and near-optimal actions.}

\todo[color=green]{Discuss that any action in the null-space of a Q function does not change the return of the agent and therefore we can allow a lower priority agent to make this choice to collect their reward.}
The general intuition of classical null-space control is also applicable to \acrshort{rl} policies. We refer to this as the \emph{indifference-space} of policies and their $Q$-functions. We define all actions $a \in \mathcal{A}$ to be in the indifference-space of a policy $\pi$ and the $Q$-function $Q^\pi$ in state $s\ \in \mathcal{S}$, if they have the same (or better) expected return as $\pi$.
In our example, the indifference-space of $\pi^*_1$ contains all actions as long as the state is not next to the obstacle, where it only contains one action.
In such a case, we relax this definition and allow some loss of return $\varepsilon$, such that all actions $a$ are part of the \textit{soft indifference-space} if 
\begin{equation}
Q^{\pi}(s, a) - Q^{\pi}(s, \pi(s)) \le \varepsilon.	
\label{eq:epsilon-constant}
\end{equation}

Intuitively, when the soft indifference-space of a high-priority policy contains multiple actions,
we can allow the low-priority policy to select which of those near-optimal actions to execute. In this way, the low-priority task can be achieved in the long run with actions that are in the high-priority policy's soft indifference space.
\todo[color=green]{Discuss that it is an open question how to do this and that you will present a possible approach to this below.}

In some tasks, this soft indifference-space might be very large (i.e.\ sparse reward navigation tasks) while in tasks with more constrained optimal policies this space might be rather small. 
In either case and to the best of our knowledge, how to find the soft indifference-space or a similar concept for \acrshort{rl} policies is an open research question, since existing \acrshort{rl} composition approaches only consider ``AND"~\cite{russell2003q, haarnoja2018composable, hunt2019composing, barreto2017successor} or ``XOR"~\cite{precup1998theoretical} task compositions. Furthermore, how to have a low-priority policy select actions from the high-priority policy's soft indifference-space, what the resulting task-prioritized compound policy or its $Q$-function is, and what this means in terms of reward are also open research questions.
In this paper, we propose a possible method for finding the soft indifference-space of a policy, for exploiting it for local task-prioritized \acrshort{rl} policy composition, and for a constrained exploration based on this space.

\todo[color=green]{Clearly state what the challenges are when we want to do this: We do not know the new common reward function. The individual optimal policies are not optimal together. We cannot just simply average the policies. There are situations where the top policy must be in total control, e.g. in a safety task situation. We must guarantee that. If the top task has preferences over actions, clearly the lower-level task need to adapt. Somehow, we need to identify when the higher priority policy is indifferent wrt. a set of actions, identify that set and let the next level policy decide between them. How is this done?}

\section{Indifference-space reinforcement learning}

\todo[color=green]{State the situation: We have several tasks within the same MDP but with different reward functions and we have priorities. We also have the optimal policies and Q functions for these tasks.}
For our task-prioritized composition framework, we consider a \acrfull{mdp} with $d$-dimensional action space  $\mathcal{A}$ and state space $\mathcal{S}$, transition dynamics $p(s_{t+1} |s_t, a_t)$, the marginal $\rho$, and a discount factor $\gamma$. 
Tasks are defined by reward functions $r\in\mathcal{R}$ for the same \acrshort{mdp} and we assume access to optimal task policies $\pi_1^*$, $\pi_2^*$ and $Q$-functions $Q_1^*$, $Q_2^*$ for tasks $r_1$, $r_2$.
There are multiple challenges to overcome: As mentioned before, we do not know the task $r_{1\succeq2}$ in terms of reward,
we cannot exclusively rely on the individual $Q$-functions for evaluating the compound task~\cite{russell2003q}, 
and, we must ensure the priorities while allowing progress on the lower-priority task when possible.

We require that $\pi_1^*$ and $\pi_2^*$ are stochastic, maximum entropy policies, meaning our method is based on the maximum-entropy framework~\cite{ziebart2008maximum, haarnoja2017reinforcement}, where policies not only maximize reward, but also the entropy $\mathcal{H}$:
\begin{equation}
\pi_\mathrm{ME}^* = \arg \underset{\pi}{\max} \sum_t \underset{(s_t, a_t) \sim \rho_\pi}{\mathbb{E}} \big[r(s_t, a_t) + \alpha \mathcal{H}(\pi(\cdot | s_t)) \big].	
\end{equation}
The maximum entropy framework is particularly adequate for policy composition, as described by \citeauthor{haarnoja2018composable}~\cite{haarnoja2018composable}, while also offering desirable mathematical properties. As we will see, the policy definition
\begin{equation}
	\pi(a_t | s_t) \propto \exp(-\frac{1}{\alpha}Q_\mathrm{Soft}(s_t, a_t)),
	\label{eq:pi_prop_q}
\end{equation}
together with \acrfull{asvgd} and the Soft $Q$-Learning~\cite{haarnoja2017reinforcement} update mechanism conveniently allow us to learn a sampling network for the intractable policy density reflecting dynamically weighted $Q$-function mixture resulting from our method.

\todo[color=green]{State what we want to achieve: We are now in a new situation where we want to solve these task together, but with priorities.}
Additionally, for prioritized composition, we must define an ordering of tasks ${1\succeq2}$, such that task priority can be determined.
Given such an ordering, we are now interested in obtaining the policy $\pi_{1\succeq2}$ for task $r_{1\succeq2}$, which maximizes both $r_1$ and $r_2$, while guaranteeing that the return for the high-priority task $r_1$ is not considerably worse under $\pi_{1\succeq2}$ compared to its optimal policy $\pi_1^*$.
\todo[color=yellow]{TSV: said differently, $r_i$ imposes a constrain on the maximization of $r_j$. Incorporate?}

\todo[color=green]{We will now explain our approach to this null-space idea from above exemplarily for the ME RL framework. Why this framework?}

\subsection{Indifference-weighting for local composition}
To obtain the task-prioritized compound policy $\pi_{1\succeq2}$, we define its $Q$-function as
\begin{equation}
Q_{1\succeq2}(s_t, a_t) = Q_1^*(s_t, a_t) + w_{1\succeq2}(s_t, a_t)Q_2^*(s_t, a_t)
\label{eq:q_comp}
\end{equation} 
and let $Q_{1\succeq2}$ induce $\pi_{1\succeq2}$ via Eq.~\eqref{eq:pi_prop_q}.
The question is how to choose $w_{1\succeq2}$ for it to implement the desired task-prioritized composition based on the high-priority soft indifference-space. 
Informally and considering our definition of the soft indifference-space of a policy, $w_{1\succeq2}(s_t, a_t)$ 
must tend to zero for all actions that are far from local maxima in the high-priority $Q$-function $Q_1^*$, 
such that the low-priority $Q$-function $Q_2^*$ can only contribute to $Q_{1\succeq2}$ at points that have similar high return under the high-priority task $r_1$. 

We can calculate such a factor by considering the first- and second-order information of the high-priority policy. The Jacobian $\mathbf{J}^{Q_1^*}$ of the high-priority policy's $Q$-function (evaluated at point $s,a$) is small around critical points, while the corresponding Hessian $\mathbf{H}^{Q_1^*}$ is negative around local maxima. 
Thus, when $Q_1^*$ is continuous and differentiable w.r.t $a_t$, we can perform a point-wise second-order derivative test to find approximate local maxima of the high-priority policy $\pi_1$, 
construct the soft indifference-space around those local maxima,
and define $w_{1\succeq2}$ as a measure of local optimality: 
\begin{equation}
	w_{1\succeq2}(s_t, a_t) \equiv \kappa(||\mathbf{J}^{Q_1}||, 0) \cdot \mathcal{F}(||\mathbf{H}^{Q_1}||).
	\label{eq:w}
\end{equation}
Here, $||.||$ denotes an arbitrary matrix norm, $\kappa$ denotes the squared-exponential (SE) kernel 
$\kappa (x, x') = \sigma^2 \mathrm{e}^{-(x-x')^2 / 2l^2}$ 
that will be large when $x$ and $x'$ are similar, while $\mathcal{F}$ denotes the Fermi-Dirac distribution 
$\mathcal{F}(\mathcal{E}) = \frac{1}{exp(\mathcal{E}-\mu / k_B T) +1}$, 
with the Boltzmann constant $k_B$ and the parameters for the point of symmetry $\mu$ and absolute temperature $T$. 
To provide further intuition for Eq.~\eqref{eq:w}, consider Fig.~\ref{fig:weight_params} which shows the two component functions of $w_{1\succeq2}$. Eq.~\eqref{eq:w} simply ensures that $w_{1\succeq2}$ tends to $1$ when the Jacobian of $Q_1$ is close to zero and the Hessian of $Q_1$ is negative, while tending to zero otherwise, which is the desired factor. However, $\kappa$ and $\mathcal{F}$ have critical parameters that control how much deviation from the optimal action $w_{1\succeq2}$ allows, which require a task dependent tuning-process that might be brittle if performed manually. It might be possible to automatically tune these parameters such that Eq.~\eqref{eq:epsilon-constant} holds for a desired $\varepsilon$, but how to do this concretely is ongoing research.

Given this definition for $w_{1\succeq2}$, we can now obtain the task-prioritized compound $Q$-function $Q_{1\succeq2}$ in Eq.~\eqref{eq:q_comp}, from two constituent $Q$-functions and a given priority ordering. $w_{1\succeq2}$ ensures that the low-priority $Q$-function can only select actions that are in the soft indifference-space of the high-priority $Q$-function. 
\todo[color=green]{Discuss only one approach in more detail (e.g. the sampling) and only mention the other ones. (Space)}
Next, we show one way for obtaining the policy that corresponds to the zero-shot, task-prioritized composition $Q$-function $Q_{1\succeq2}$ and note that this approach is just one of potentially many ways for implementing task-prioritized composition in \acrshort{rl}.

\subsection{Learning the task-prioritized compound policy}
Given $Q_{1\succeq2}$ from above, we must still learn a sampling model for the corresponding, potentially high-dimensional and intractable density of the task-prioritized compound maximum entropy policy $\pi_{1\succeq2}$. We rely on the \acrshort{asvgd}~\cite{feng2017learning}, as in original Soft Q-learning~\cite{haarnoja2017reinforcement}, to train a \acrshort{dnn} $f$ parameterized by $\phi$ to map random noise inputs $\zeta$ to action samples $a_t$ from the maximum entropy target policy $\pi_{1\succeq2}$ such that $a_t = f^\phi(\zeta, s_t)$. 
The Stein variational gradient that minimizes the Kullback-Leibler divergence between the policy $\pi^\phi$ (induced by $f^\phi$) and the desired, intractable policy density $\pi_{1\succeq2}$ corresponding to the soft $Q$-function $Q_{1\succeq2}$ is given by 
\begin{multline*}
\nabla f^\phi(\cdot,  \mathbf{s}_t) = \mathbb{E}_{\mathbf{a}_t \sim \pi^\phi} \big [ \kappa(\mathbf{a}_t, f^\phi(\zeta, \mathbf{s}_t)) \nabla \underbrace{Q_{1\succeq2}(\mathbf{s}_t, \mathbf{a}_t)}_{\propto\pi_{1\succeq2}} \\ + \alpha \nabla k(f^\phi(\zeta, \mathbf{s}_t)) \big],
\end{multline*}
where $\kappa$ is an arbitrary kernel.
This gradient can directly be used to update the sampling network (for details see \cite{haarnoja2017reinforcement, feng2017learning}). The reason this works is the proportion in Eq.~\eqref{eq:pi_prop_q}, which implies that the soft $Q$-function $Q_{1\succeq2}$ corresponds to the un-normalized density of $\pi_{1\succeq2}$, hence making the \acrshort{asvgd} method applicable.
Thus, as long as we can calculate $Q_{1\succeq2}(s_t, a_t)$ for a batch of states and actions, we can train a sampling network for the corresponding intractable policy $\pi_{1\succeq2}$. 
In this fashion, we can directly use the soft $Q$-learning \acrshort{asvgd} update procedure to learn sampling networks for high-dimensional, task-prioritized composition policy densities induced by indifference-weighted $Q$-functions.

\todo[color=green]{Discuss the limitation of local combination. Clearly, we can get stuck in local optimal of Q functions and clearly, the produced policy is not optimal. We can even show in a simple example that it does get stuck in the U and that a better and priority aware policy exists. That was also the case in null-space control. But in RL, we can do something about this. We can continue to learn, starting with this policy we generated. This should be much more efficient and it should be safe.}
The procedure outlined so far is a semi-direct adaptation of null-space control for \acrshort{rl} policies and $Q$-functions, meaning it shares the limitations mentioned in Sec.~\ref{sec:null-space-control}, 
namely that it only yields a locally optimal policy for the task $r_{1\succeq2}$ that is optimal for $r_1$ but not necessarily for $r_2$, which can lead to the locally optimal behavior sketched in Fig.~\ref{fig:env}.
With a classical null-space controller, this would be the resulting, locally optimal behavior, with no way of improving it, except by manually programming a low-priority controller that optimally solves the low-priority task in the null-space of the high-priority controller. 
However, a globally optimal solution clearly exists, namely driving around the obstacle as seen in Fig.~\ref{fig:env} on the right. Our indifference-space composition approach for \acrshort{rl} can learn this globally optimal policy for the new compound task $r_{1\succeq2}$, as we explain in the next section.

\subsection{Online learning with indifference-space exploration}

\todo[color=green]{Here you could continue with the U example with only two policies. Only the second policy needs to learn because the first one is the top policy. Can we sample in an informed way? Can we explore without inflicting loss on the top policy? Can we demonstrate that we can make an algorithm that finds a better policy that does not get stuck?} 
Given our procedure for finding the soft indifference-space of a $Q$-function, while keeping the high-priority $Q$-function $Q_1^*$ fixed, we can continue to train the low-priority $Q$-function $Q_2^*$. By sampling exploratory actions in the soft indifference-space of the high-priority policy, the low-priority policy can learn to adapt to the constraints imposed by the high-priority policy. 
The reason we can do this while still guaranteeing constrain-satisfaction for the high-priority task is that the indifference weight $w_{1\succeq2}$ will filter out the contribution from the low-priority $Q$-function $Q_2$ at all points that are far from local maxima in $Q_1^*$, such that even an adversarial low-priority $Q$-function that maximizes the task $-r_1$ opposite to $r_1$ would not result in constrain violation for the high-priority task $r_1$. 
Thus, while keeping the parameters of $Q_1^*$ frozen, we can take exploratory actions in the soft indifference-space of $Q_1^*$ to learn a new state-action value function $Q_2^\sim$ (possibly with initial parameters from $Q_2^*$) that adapts to the constraints imposed by $Q_1^*$. Thus, while Eq.~\eqref{eq:q_comp} can be thought of as the equivalent to locally optimal null-space control in \acrshort{rl}, with this indifference-space exploration strategy we can obtain the globally optimal policy $\pi_{1\succeq2}^*$ induced by
\begin{equation}
	Q_{1\succeq2}^*(s_t, a_t) = Q_1^*(s_t, a_t) + w_{1\succeq2}(s_t, a_t)Q_2^\sim(s_t, a_t),
	\label{eq:optimal_q_comp}
\end{equation}
as described in the previous section. $\pi_{1\succeq2}^*$ will be globally optimal for the task $r_{1\succeq2}$, meaning both tasks will be maximized jointly, in the best possible way under the task priority ordering. In Fig.~\ref{fig:env}, this policy will produce the desired, globally optimal behavior in which the agent navigates around the U-obstacle instead of down the middle.
\linebreak \linebreak
We aim to present initial experimental results of our method at the 2nd RL-CONFORM workshop at IROS 2022.
  

\todo{I want a concluding statement here but don't think ``conclusion'' would be adequate, because we can't really conclude anything without experimental results. How to call this section instead? }

\addtolength{\textheight}{-12cm}   


\printbibliography

\end{document}